\documentclass{article}

\usepackage[preprint]{timeseries_workshop}

\usepackage[utf8]{inputenc} 
\usepackage[T1]{fontenc}    
\usepackage{hyperref}       
\usepackage{url}            
\usepackage{booktabs}       
\usepackage{amsfonts}       
\usepackage{nicefrac}       
\usepackage{microtype}      

\usepackage{tabularx}
\usepackage{multirow}
\usepackage{array}
\usepackage{caption}
\usepackage{graphicx} 
\usepackage{colortbl} 
\usepackage{amsmath}  
\usepackage{subcaption}
\usepackage{hyperref}
  \hypersetup{
    final=true,
    plainpages=false,
    pdfstartview=FitV,
    pdftoolbar=true,
    pdfmenubar=true,
    bookmarksopen=true,
    bookmarksnumbered=true,
    breaklinks=true,
    linktocpage=true,
    colorlinks=true,
    linkcolor=red, 
    urlcolor=iris, 
    citecolor=darkcerulean, 
    anchorcolor=black
  }

\usepackage[table]{xcolor}
\definecolor{darkcerulean}{rgb}{0.03, 0.27, 0.49}

\usepackage{mymath}         

\title{User-friendly Foundation Model Adapters for Multivariate Time Series Classification}

\author{%
  Vasilii Feofanov$^{*1}$ \quad Romain Ilbert\thanks{Equal Contribution}$^{*1,2}$ \quad    Malik Tiomoko$^1$  \quad \\ \textbf{Themis Palpanas}$^2$ \quad \textbf{Ievgen Redko}$^1$ \\
  $^1$Huawei Noah’s Ark Lab, Paris, France \quad $^2$LIPADE, Paris Descartes University, Paris, France 
}

\begin{document}

\maketitle

\begin{abstract}
Foundation models, while highly effective, are often resource-intensive, requiring substantial inference time and memory. This paper addresses the challenge of making these models more accessible with limited computational resources by exploring dimensionality reduction techniques. Our goal is to enable users to run large pre-trained foundation models on standard GPUs without sacrificing performance. We investigate classical methods such as Principal Component Analysis alongside neural network-based adapters, aiming to reduce the dimensionality of multivariate time series data while preserving key features. Our experiments show up to a 10x speedup compared to the baseline model, without performance degradation, and enable up to 4.5x more datasets to fit on a single GPU, paving the way for more user-friendly and scalable foundation models.
\end{abstract}

\section{Introduction}

The remarkable success of pre-trained models in natural language processing (NLP) \citep{achiam2023gpt,touvron2023llama} and computer vision \citep{dosovitskiy2021vit} has inspired the extension of this paradigm to time series data. Time Series Foundation Models (TSFMs) aim to generalize across diverse downstream tasks by learning versatile encoders from large, heterogeneous pre-training datasets. This strategy offers both flexibility and efficiency, as deploying TSFMs for new tasks requires only fine-tuning, thus reducing the reliance on extensive labeled training data.

Depending on their pre-training objectives, TSFMs can be specialized for tasks like forecasting \citep{garza2023timegpt,rasul2023lagllama,wang2024rose}, classification \citep{lin2024nutime}, or designed to tackle various time series problems \citep{zhou2023onefitsall,goswami2024moment}. However, most existing models are univariate, necessitating separate applications to each channel in multivariate data. This approach poses significant limitations when dealing with datasets that have hundreds or thousands of channels \citep{wei2018multivariatebook,bagnall2018uea}, leading to increased runtime and memory consumption, especially when fine-tuning on limited computational resources.

In this paper, we address this overlooked challenge by integrating dimensionality reduction techniques with foundation models for multivariate time series analysis. While dimensionality reduction \citep{van2009dimensionality} and feature selection \citep{guyon2003introduction} are well-established individually, their combination with foundation models introduces unique challenges and hidden obstacles. We explore various methods, including Principal Component Analysis (PCA) and neural network-based adapters, to preprocess multivariate data and alleviate computational and memory constraints.

Our experiments demonstrate up to a 10x speedup and enable up to 4.5x more datasets to fit on a single GPU, all while maintaining classification accuracy, as verified by pairwise p-value tests. These results highlight the potential of dimensionality reduction to make foundation models more efficient and accessible for multivariate time series classification.

\section{Related Work}
Classical models for time series classification, including those based on Dynamic Time Warping~\citep{salvador2007toward,cuturi2017soft}, kernel methods~\citep{salvador2007toward,cuturi2017soft}, shapelet-based algorithms~\citep{lines2012shapelet}, tree-based models~\citep{deng2013time}, and dictionary-based approaches~\citep{lin2007experiencing,lin2012rotation}, are effective for univariate time series but face challenges when extended to multivariate time series (MTS). Deep learning methods and random convolution techniques like ROCKET~\citep{dempster2020rocket} and Multi-ROCKET show promise but typically treat each channel independently, leading to scalability and computational issues. TSFMs~\citep{goswami2024moment, wang2024rose, garza2023timegpt, zhou2023onefitsall, rasul2023lagllama}, inspired by advances in NLP and computer vision, offer potential for MTS classification but still struggle with complexity and inter-channel dependencies.

\section{Framework}
\subsection{Problem setup}

\paragraph{Notations.} Let $N$ represent the number of samples, $T$ the number of time steps, $D$ the number of channels or dimensions in each multivariate time series, and $D'$ the reduced number of dimensions after applying dimensionality reduction, with $D' \leq D$.

\paragraph{Datasets.} We use 12 multivariate datasets from the UEA repository~\citep{bagnall2018uea}, each with at least 10 channels to enable meaningful dimensionality reduction. Detailed dataset characteristics are provided in Appendix~\ref{app:datasets}.

\paragraph{Experimental Setup.} All experiments were performed on a single NVIDIA Tesla V100-32GB GPU, with a 2-hour limit per run. Runs exceeding this limit are marked \texttt{TO} (Time Out), while those facing CUDA out-of-memory issues are labeled \texttt{COM} (CUDA Out of Memory).

\paragraph{Foundation Models.} We evaluate two TSFMs: \texttt{MOMENT}, a large-scale model with 341 million parameters~\citep{goswami2024moment}, and \texttt{ViT}, a smaller model with 8 million parameters, inspired by ViT-based models like Nu-Time~\citep{lin2024nutime} and PatchTST~\citep{nie2022patchtst}. More implementation details are provided in Appendix~\ref{app:fm}.

\paragraph{Objective.} Our goal is efficient multivariate time series classification using pre-trained models, with accuracy as the primary metric. We focus on rapid fine-tuning within a 2-hour window on a single GPU, without significant performance loss. To achieve this, we test various dimensionality reduction techniques—such as PCA and neural network-based adapters—integrated at the beginning of the foundation model pipeline, and evaluate different fine-tuning strategies.

\subsection{Motivation}
Table \ref{tab:motivation} presents the accuracy results of two TSFMs, \texttt{ViT} and \texttt{MOMENT}, on a range of multivariate time series datasets under full fine-tuning without the use of any adapter, i.e., without dimensionality reduction. Notably, the results indicate that most of the foundation models encounter severe computational limitations when applied to multivariate data on standard hardware (NVIDIA Tesla V100-32GB GPU), as indicated by the \texttt{COM} and \texttt{TO} entries. These computational constraints underscore the difficulty of directly applying existing foundation models to multivariate time series with numerous channels, often leading to excessive resource consumption and failures to complete the fine-tuning process. This evidence motivates our exploration of dimensionality reduction techniques, which aim to alleviate these computational bottlenecks and enable foundation models to handle multivariate data more effectively without compromising accuracy.

\begin{table}[ht!]

\caption{Accuracy averaged over 3 model runs when the models are under full fine-tuning without an adapter (i.e., using all initial channels).}
\label{tab:motivation}
{\scalebox{0.65}{
\begin{tabular}{l|llllllllllll}
\toprule
Model          & Duck & Face & Finger & Hand & Heart & Insect & Vowels & Motor & NATOPS            & PEMS & Phoneme    & SpokeA \\
\midrule
\texttt{ViT} & \texttt{COM}           & \texttt{COM}           & \texttt{COM}             & .401 $\pm$ .021        & \texttt{COM}       & \texttt{COM}                  & .981 $\pm$ .005   & \texttt{COM}          & .937 $\pm$ .012    & \texttt{COM}     & .342 $\pm$ .002    & .987 $\pm$ .001     \\
\texttt{MOMENT}          & \texttt{COM}           & \texttt{COM}           & \texttt{COM}             & .356 +- .016 & \texttt{COM}       & \texttt{COM}                  & .925 +- .002 & \texttt{COM}          & \texttt{TO} & \texttt{COM}     & \texttt{TO} & \texttt{TO}\\
\bottomrule
\end{tabular}
}}
\end{table}

\subsection{Feature-Level Transformation Methods}

We explore several dimensionality reduction techniques to preprocess multivariate time series data for foundation models.

\paragraph{Principal Component Analysis (PCA)} seeks to find an orthogonal basis of principal components where a few components capture most of the data's variance. Applying PCA to 3D matrices $(N, T, D)$ poses challenges. A common approach reshapes the data into $(N, T \times D)$ and projects it to $(N, T \times D')$, but this disrupts the temporal structure. Additionally, when $N \ll T \times D$, PCA can become computationally unstable. To address this, we reshape the data to $(N \times T, D)$, allowing PCA to focus on correlations between channels over all time steps, effectively capturing spatial correlations while preserving temporal information. The learned rotation matrix $W \in \mathbb{R}^{D' \times D}$ linearly combines the original channels into a lower-dimensional space, applied consistently across all time steps.

\paragraph{Truncated Singular Value Decomposition (SVD)} also reduces dimensionality by retaining the most significant components. Unlike PCA, SVD operates directly on the data matrix without centering it, decomposing it into its top $k$ singular values and vectors. This method effectively captures the principal directions of variance.

\paragraph{Random Projection (Rand Proj)} is a computationally efficient technique that projects the data onto a lower-dimensional subspace using randomly generated directions. Unlike PCA, it does not aim to capture the most variance but instead focuses on providing a quick dimensionality reduction solution with minimal computational cost.

\paragraph{Variance-Based Feature Selection (VAR)} is a simple but effective method that selects features with the highest variance. Features with low variance are considered less informative and can be discarded without significantly affecting the overall representation of the data.

\paragraph{Linear Combiner (lcomb)} introduces a learnable adapter that performs a linear combination of channels before passing the data to the encoder and classification head. In contrast to unsupervised methods like PCA, this approach learns the rotation matrix $W \in \mathbb{R}^{D' \times D}$ in a supervised manner, either by fine-tuning the adapter and head or the entire network. Given the large search space for possible linear combinations, we apply a top-k rule to each row of $W$, retaining only the top $k$ entries to ensure more efficient optimization.

\section{Experimental Results}
\label{sec:results}

\setlength{\tabcolsep}{0.3em}
\begin{table}[ht!]
\centering
\caption{Performance comparison between different adapter configurations for \texttt{MOMENT} and \texttt{ViT} foundation models when the new number of channels is fixed to 5. Best performance is shown in \textbf{bold} and second best in \textit{italic}. Results for fine-tuning the head only are given for the reference.}
\label{tab:moment_vit_results}
\scalebox{0.6}{
    \begin{tabular}{c|c|c||ccccccc}
    \toprule
    \multirow{2}{*}{Dataset} & \multirow{2}{*}{Model} & head & \multicolumn{6}{c}{adapter+head} \\
    \cmidrule(r{10pt}l{5pt}){4-9}
    & & no adapter & PCA & SVD & Rand\_Proj & VAR & lcomb & lcomb\_top\_k \\

    \midrule
    \multirow{2}{*}{DuckDuckGeese}
    & MOMENT & $0.460_{\pm 0.016}$ & $\textit{0.627}_{\pm 0.023}$ & $\textbf{0.667}_{\pm 0.012}$ & $0.500_{\pm 0.040}$ & $0.407_{\pm 0.012}$ & $0.427_{\pm 0.046}$ & $0.393_{\pm 0.114}$ \\
    & ViT & $0.420_{\pm 0.020}$ & $\textit{0.558}_{\pm 0.023}$ & $\textbf{0.600}_{\pm 0.032}$ & $0.487_{\pm 0.023}$ & $0.400_{\pm 0.060}$ & $0.360_{\pm 0.020}$ & $0.393_{\pm 0.031}$ \\
    \midrule
    \multirow{2}{*}{FaceDetection}
    & MOMENT & $0.623_{\pm 0.006}$ & $\textbf{0.567}_{\pm 0.002}$ & $\textit{0.566}_{\pm 0.001}$ & $0.552_{\pm 0.014}$ & $0.555_{\pm 0.001}$ & TO & TO \\
    & ViT & $0.595_{\pm 0.004}$ & $\textbf{0.554}_{\pm 0.001}$ & $\textit{0.551}_{\pm 0.007}$ & $0.533_{\pm 0.004}$ & $0.539_{\pm 0.007}$ & $0.548_{\pm 0.008}$ & $0.550_{\pm 0.008}$ \\
    \midrule
    \multirow{2}{*}{FingerMovement}
    & MOMENT & $0.573_{\pm 0.012}$ & $\textit{0.593}_{\pm 0.032}$ & $0.573_{\pm 0.012}$ & $0.573_{\pm 0.025}$ & $\textbf{0.613}_{\pm 0.021}$ & $0.573_{\pm 0.032}$ & $0.540_{\pm 0.017}$ \\
    & ViT & $0.627_{\pm 0.015}$ & $\textbf{0.593}_{\pm 0.044}$ & $0.530_{\pm 0.030}$ & $0.570_{\pm 0.075}$ & $\textit{0.582}_{\pm 0.040}$ & $0.580_{\pm 0.020}$ & $0.567_{\pm 0.046}$ \\
    \midrule
    \multirow{2}{*}{HandMovementDirection}
    & MOMENT & $0.401_{\pm 0.008}$ & $\textit{0.410}_{\pm 0.043}$ & $0.365_{\pm 0.036}$ & $0.405_{\pm 0.041}$ & $0.369_{\pm 0.039}$ & $0.378_{\pm 0.047}$ & $\textbf{0.414}_{\pm 0.008}$ \\
    & ViT & $0.342_{\pm 0.021}$ & $\textbf{0.396}_{\pm 0.021}$ & $\textit{0.351}_{\pm 0.089}$ & $0.329_{\pm 0.083}$ & $0.329_{\pm 0.031}$ & $0.320_{\pm 0.034}$ & $0.320_{\pm 0.028}$ \\
    \midrule
    \multirow{2}{*}{Heartbeat}
    & MOMENT & $0.740_{\pm 0.003}$ & $0.732_{\pm 0.000}$ & $0.732_{\pm 0.005}$ & $\textbf{0.756}_{\pm 0.005}$ & $0.725_{\pm 0.006}$ & $\textit{0.737}_{\pm 0.005}$ & $\textit{0.737}_{\pm 0.013}$ \\
    & ViT & $0.811_{\pm 0.010}$ & $0.766_{\pm 0.005}$ & $0.737_{\pm 0.012}$ & $0.776_{\pm 0.013}$ & $\textbf{0.780}_{\pm 0.010}$ & $0.748_{\pm 0.006}$ & $\textit{0.779}_{\pm 0.014}$ \\
    \midrule
    \multirow{2}{*}{InsectWingbeat}
    & MOMENT & $0.284_{\pm 0.003}$ & $\textbf{0.239}_{\pm 0.003}$ & $\textit{0.224}_{\pm 0.003}$ & $0.193_{\pm 0.027}$ & $0.195_{\pm 0.004}$ & $0.167_{\pm 0.014}$ & $0.213_{\pm 0.010}$ \\
    & ViT & $0.614_{\pm 0.005}$ & $0.344_{\pm 0.013}$ & $\textit{0.352}_{\pm 0.010}$ & $0.333_{\pm 0.035}$ & $0.238_{\pm 0.012}$ & $0.171_{\pm 0.013}$ & $\textbf{0.354}_{\pm 0.041}$ \\
    \midrule
    \multirow{2}{*}{JapaneseVowels}
    & MOMENT & $0.885_{\pm 0.002}$ & $0.801_{\pm 0.009}$ & $\textit{0.803}_{\pm 0.003}$ & $0.796_{\pm 0.011}$ & $0.734_{\pm 0.008}$ & $0.797_{\pm 0.035}$ & $\textbf{0.819}_{\pm 0.027}$ \\
    & ViT & $0.979_{\pm 0.006}$ & $\textbf{0.922}_{\pm 0.009}$ & $0.897_{\pm 0.012}$ & $\textit{0.902}_{\pm 0.008}$ & $0.885_{\pm 0.010}$ & $0.798_{\pm 0.070}$ & $0.816_{\pm 0.027}$ \\
    \midrule
    \multirow{2}{*}{MotorImagery}
    & MOMENT & $0.643_{\pm 0.015}$ & $0.590_{\pm 0.010}$ & $\textbf{0.607}_{\pm 0.012}$ & $0.567_{\pm 0.032}$ & $0.550_{\pm 0.010}$ & $0.583_{\pm 0.015}$ & $\textit{0.593}_{\pm 0.025}$ \\
    & ViT & $0.600_{\pm 0.036}$ & $\textit{0.593}_{\pm 0.025}$ & $0.590_{\pm 0.017}$ & $0.577_{\pm 0.029}$ & $\textbf{0.607}_{\pm 0.025}$ & $0.557_{\pm 0.045}$ & $\textbf{0.607}_{\pm 0.055}$ \\
    \midrule
    \multirow{2}{*}{NATOPS}
    & MOMENT & $0.872_{\pm 0.011}$ & $\textit{0.776}_{\pm 0.008}$ & $0.739_{\pm 0.017}$ & $0.774_{\pm 0.032}$ & $\textbf{0.813}_{\pm 0.020}$ & $0.596_{\pm 0.017}$ & $0.769_{\pm 0.031}$ \\
    & ViT & $0.944_{\pm 0.011}$ & $\textbf{0.874}_{\pm 0.014}$ & $0.820_{\pm 0.012}$ & $\textit{0.852}_{\pm 0.038}$ & $0.850_{\pm 0.035}$ & $0.787_{\pm 0.003}$ & $0.826_{\pm 0.036}$ \\
    \midrule
    \multirow{2}{*}{PEMS-SF}
    & MOMENT & $0.834_{\pm 0.026}$ & $0.678_{\pm 0.007}$ & $0.511_{\pm 0.022}$ & $0.644_{\pm 0.027}$ & $0.611_{\pm 0.015}$ & $\textbf{0.740}_{\pm 0.010}$ & $\textit{0.697}_{\pm 0.013}$ \\
    & ViT & $0.923_{\pm 0.023}$ & $\textbf{0.674}_{\pm 0.032}$ & $\textit{0.640}_{\pm 0.045}$ & $0.615_{\pm 0.023}$ & $0.615_{\pm 0.055}$ & $0.584_{\pm 0.025}$ & $0.594_{\pm 0.065}$ \\
    \midrule
    \multirow{2}{*}{PhonemeSpectra}
    & MOMENT & $0.234_{\pm 0.001}$ & $\textit{0.234}_{\pm 0.002}$ & $0.212_{\pm 0.002}$ & $\textbf{0.245}_{\pm 0.003}$ & $0.228_{\pm 0.004}$ & TO & TO \\
    & ViT & $0.296_{\pm 0.003}$ & $0.270_{\pm 0.003}$ & $0.259_{\pm 0.001}$ & $\textit{0.293}_{\pm 0.002}$ & $\textbf{0.294}_{\pm 0.004}$ & $0.279_{\pm 0.002}$ & $0.286_{\pm 0.001}$ \\
    \midrule
    \multirow{2}{*}{SpokenArabicDigits}
    & MOMENT & $0.977_{\pm 0.001}$ & $\textit{0.972}_{\pm 0.000}$ & $\textbf{0.978}_{\pm 0.000}$ & $0.961_{\pm 0.008}$ & $0.935_{\pm 0.002}$ & TO & TO \\
    & ViT & $0.940_{\pm 0.003}$ & $\textbf{0.962}_{\pm 0.003}$ & $0.933_{\pm 0.001}$ & $0.879_{\pm 0.004}$ & $\textit{0.946}_{\pm 0.003}$ & $0.834_{\pm 0.019}$ & $0.873_{\pm 0.019}$ \\

    \bottomrule
    \end{tabular}
}
\end{table}

\begin{figure}[ht!]
    \centering
    \begin{subfigure}[b]{0.5\textwidth}
        \centering
        \includegraphics[scale=0.2]{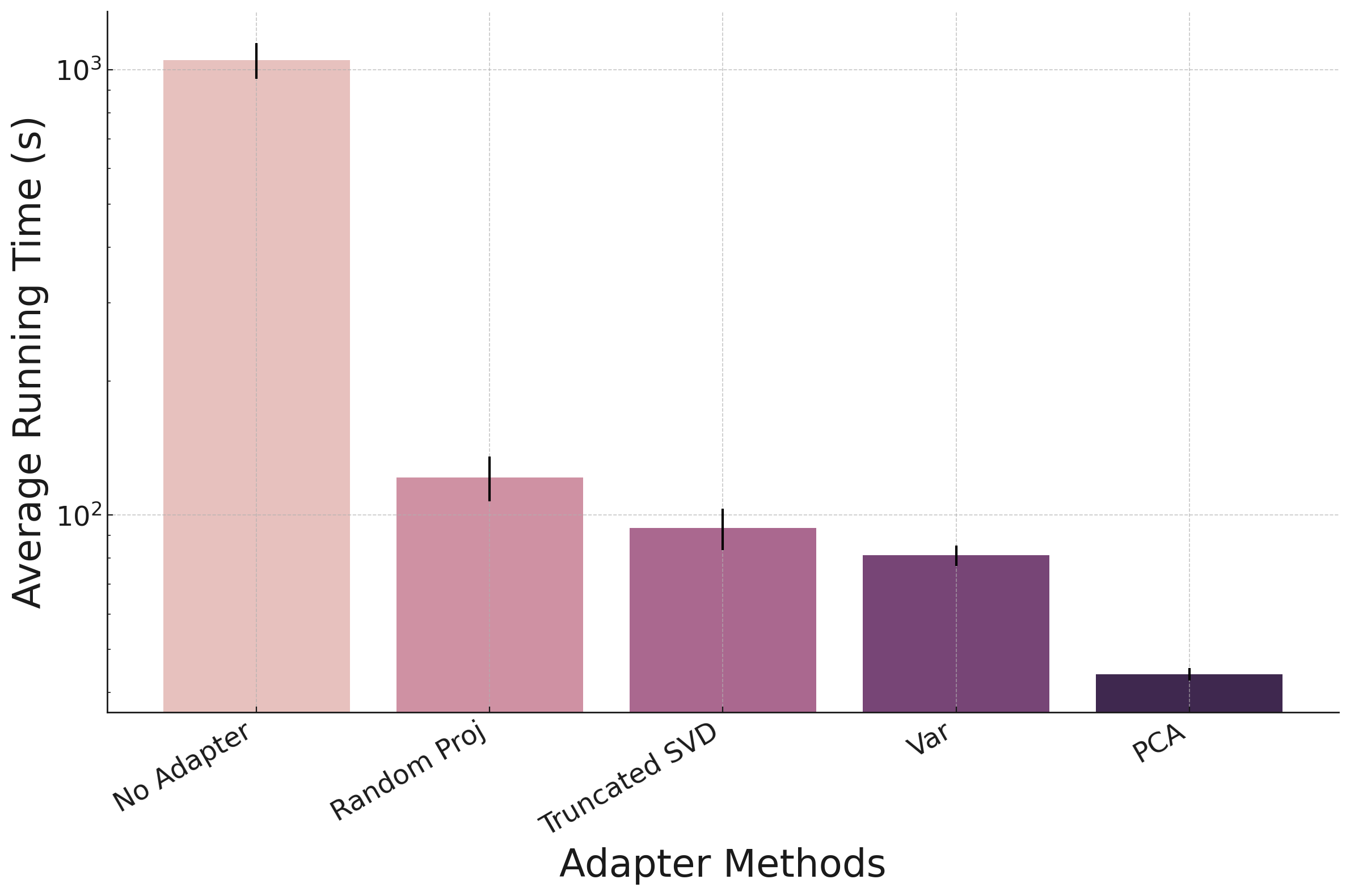}  
        \caption{Running Time for \texttt{MOMENT} Foundation Model}
        \label{fig:moment_running_time}
    \end{subfigure}%
    \hfill
    \begin{subfigure}[b]{0.5\textwidth}
        \centering
        \includegraphics[scale=0.2]{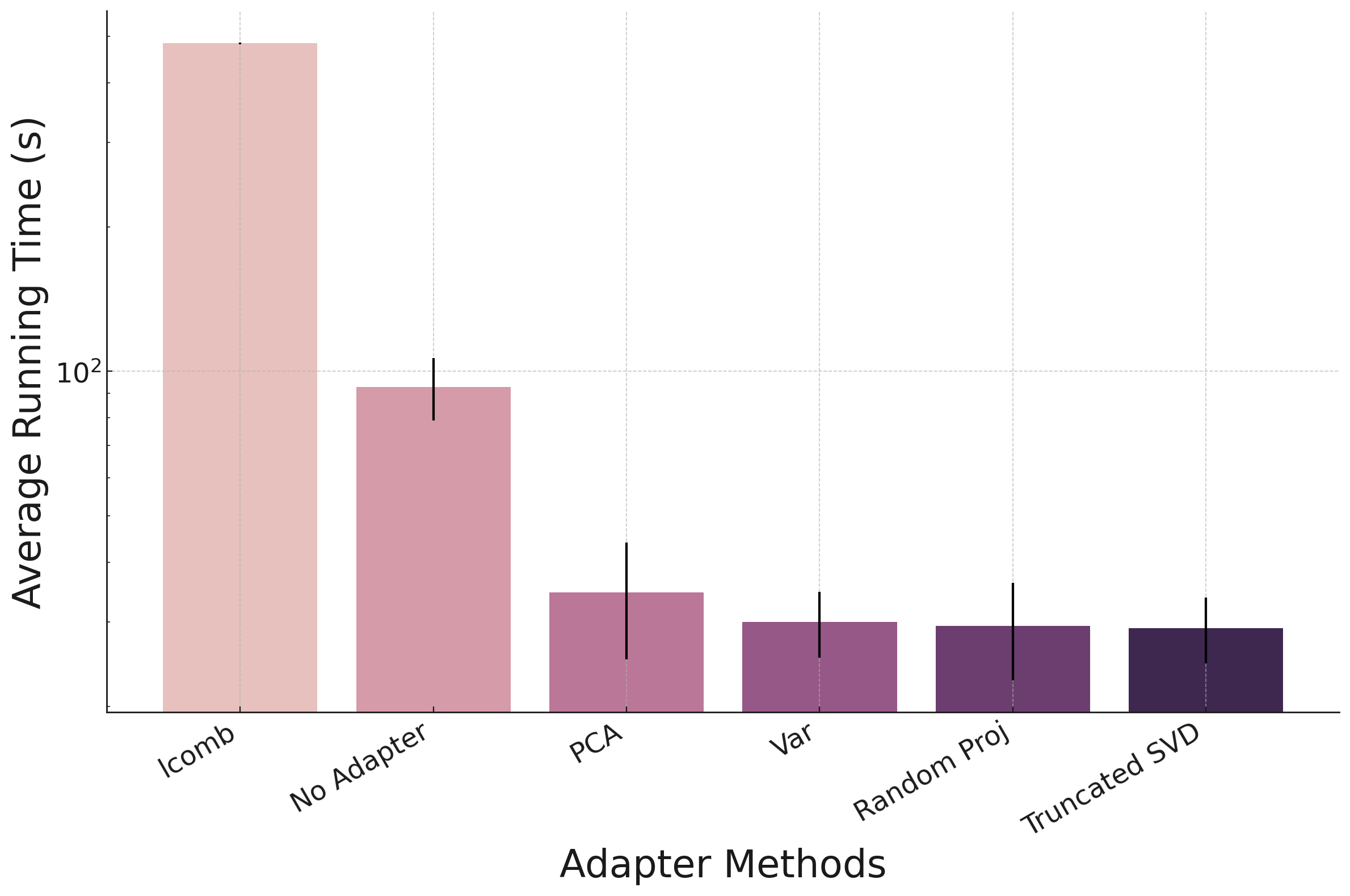}  
        \caption{Running Time for ViT Foundation Model}
        \label{fig:ViT_running_time}
    \end{subfigure}
    \caption{Comparison of running times for \texttt{MOMENT} and ViT Foundation Models averaged across all datasets and three different seeds}
    \label{fig:comparison_ViT_moment}
\end{figure}



We present the experimental comparison between different adapters when fine-tuning both the adapter and the head a foundation model.
The head refers to a classification linear layer at the end of the model, while the adapter is inserted before the foundation model. We report results for \texttt{MOMENT} and \texttt{ViT} across twelve datasets from the UEA archive with more than ten features (see Appendix \ref{app:datasets} for more details), reducing dimensionality to five channels. Also, we report the results when fine-tuning the head without an adapter.

The results, presented in Table~\ref{tab:moment_vit_results}, along with statistical tests in Appendix~\ref{app:stats}, show no statistically significant difference between the method in average over all datasets, including fine-tuning the head only. However, as shown in Figure~\ref{fig:comparison_ViT_moment} , using adapters significantly reduces computation time. For instance, with \texttt{MOMENT}, adapters are on average over ten times faster than without adapters, and for \texttt{ViT}, they provide a two-fold speed increase. 

The exception is the Linear Combiner (\texttt{lcomb}) adapter, a deep learning-based model requiring training and inference on the foundation model at every fine-tuning step. In contrast, other non-deep learning adapters process the data once to generate embeddings, allowing inference and fine-tuning of the head only, without repeatedly running the foundation model. This substantially reduces computation time compared to methods like \texttt{lcomb}.

In Table~\ref{tab:moment_vit_results}, we can see that the no adapter approach outperforms on some specific datasets, which indicates that the intrinsic dimension is dataset-dependent and there is need in more complex adapter configurations to achieve sparse dimension reduction in the general case.

By comparing the results in Appendix~\ref{app:full} with those in Table~\ref{tab:motivation}
, we observe that with the \texttt{lcomb} method, for example, we can now fine-tune 12 out of 12 datasets for \texttt{ViT} and 9 out of 12 datasets for \texttt{MOMENT} on a single GPU, compared to previously only 5 and 2 datasets, respectively, for full fine-tuning. This represents 2.4x more datasets that fit on a single GPU in less than two hours for \texttt{ViT} and 4.5x more for \texttt{MOMENT}.

\section{Conclusion}

We addressed computational and memory challenges in fine-tuning foundation models for multivariate time series by introducing dimensionality reduction techniques. These methods significantly improved efficiency, achieving up to 10x faster fine-tuning and enabling up to 4.5x more datasets to fit on a single GPU, while maintaining comparable performance. Our results highlight the potential of adapters to enhance the scalability of foundation models. Future work may focus on further optimizing these techniques and applying them to larger datasets and more complex time series tasks.

\bibliographystyle{apalike}
\bibliography{bibliography}



\appendix

\section{Experimental setup}
\subsection{Datasets}
\label{app:datasets}
The experimental results presented in this work are based on a diverse set of datasets, whose main characteristics are summarized in Table \ref{tab:datasets}. These datasets span a variety of domains and tasks, offering a comprehensive evaluation of the fine-tuning methods under consideration. For instance, the datasets include time-series data from physiological measurements (e.g., \textit{Heartbeat}, \textit{MotorImagery}), sensor readings (e.g., \textit{PEMS-SF}), and acoustic signals (e.g., \textit{PhonemeSpectra}, \textit{SpokenArabicDigits}). The number of channels, sequence lengths, and class distributions vary significantly across datasets, ensuring that the results generalize across different data modalities and problem settings.

In the case of the \textit{InsectWingbeat} dataset, we specifically subsampled 1000 examples from the original training set (which contains 30,000 examples) and 1000 from the original test set (of 20,000 examples) to reduce computational overhead while maintaining sufficient variety in the data for robust model evaluation. Each dataset was carefully chosen to challenge the models across different feature spaces, class imbalances, and temporal dependencies. For example, the \textit{JapaneseVowels} dataset focuses on speaker classification based on vowel sounds, while the \textit{DuckDuckGeese} dataset involves distinguishing animal sounds with varying levels of complexity in terms of sequence length and channel dimensionality.

By including these datasets, we ensure that the evaluation framework captures the performance of fine-tuning methods across a wide spectrum of classification tasks.


\begin{table}[ht!]
\centering
\caption{Main characteristics of the considered datasets.}
\scalebox{0.75}{
\begin{tabular}{l|ccccc}
\toprule
Dataset                          & Train Size & Test Size & \# of channels & Sequence Len & \# of classes \\
\midrule
DuckDuckGeese (Duck)             & 60         & 40        & 1345           & 270          & 5             \\
FaceDetection (Face)             & 5890       & 3524      & 144            & 62           & 2             \\
FingerMovements (Finger)         & 316        & 100       & 28             & 50           & 2             \\
HandMovementDirection (Hand)     & 320        & 147       & 10             & 400          & 4             \\
Heartbeat (Heart)                & 204        & 205       & 61             & 405          & 2             \\
InsectWingbeat (Insect)          & 1000       & 1000      & 200            & 78           & 10            \\
JapaneseVowels (Vowels)       & 270        & 370       & 12             & 29           & 9             \\
MotorImagery (Motor)             & 278        & 100       & 64             & 3000         & 2             \\
NATOPS                           & 180        & 180       & 24             & 51           & 6             \\
PEMS-SF (PEMS)                   & 267        & 173       & 963            & 144          & 7             \\
PhonemeSpectra (Phoneme)         & 3315       & 3353      & 11             & 217          & 39            \\
SpokenArabicDigits (SpokeA) & 6599       & 2199      & 13             & 93           & 10      \\     
\bottomrule
\end{tabular}
}
\label{tab:datasets}
\end{table}

\section{Implementation Details}
\subsection{Foundation Models}
\label{app:fm}

For the \texttt{MOMENT} model, we utilized the HuggingFace checkpoint provided by the authors \citep{goswami2024moment}. In contrast, for \texttt{ViT}, we implemented and trained the model ourselves, initially aiming to replicate the Nu-Time architecture \citep{lin2024nutime}, as the source code is currently unavailable. However, since we were unable to achieve comparable experimental results, our implementation diverges in certain aspects. Specifically, we extract overlapping patches from the time series, which are further embedded with statistical embeddings to form tokens that are processed by a transformer. During training, we employ a variant of the InfoNCE loss \citep{oord2018representation} proposed by \citet{he2020momentum}.

\section{Experimental Details}

\subsection{PCA's Hyperparameter Sensitivity}

\begin{figure}[h!]
    \centering
    \includegraphics[scale=0.35]{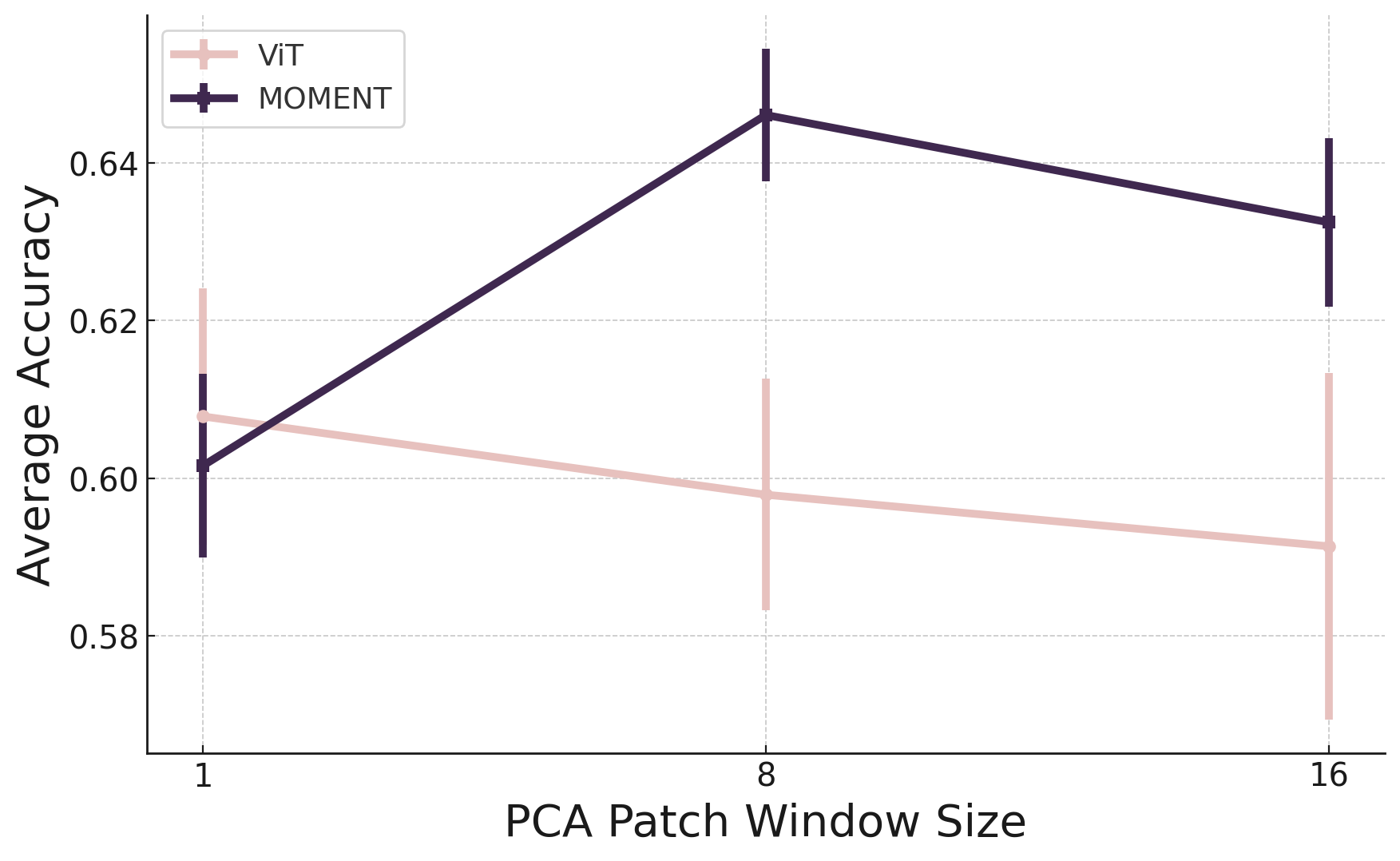}  
    \caption{Comparison of PCA and PatchPCA Methods for \texttt{ViT} and \texttt{MOMENT} Models}
    \label{fig:patchpca}
\end{figure}

In this experiment, we implemented a variant of PCA called Patch PCA. Unlike the traditional approach where the input time series of shape $(N, T, D)$ is reshaped into $(N \times T, D)$ before applying PCA, our method reshapes the input into $(N \times n_p, pws \times D)$, where $n_p$ represents the number of patches in the sequence and $pws$ refers to the patch window size. The case where $pws = 1$ corresponds to the standard PCA approach. We compare the results across different patch window sizes ($pws = 1, 8, 16$), as seen in Figure \ref{fig:patchpca}. These experiments show no clear pattern in performance across the different patch sizes, suggesting that the patch window size can be treated as a hyperparameter to be tuned based on the specific dataset.

Furthermore, we introduced two key hyperparameters for our PCA implementation: the patch window size ($pws$) and the option to scale the data before performing PCA. The results of PCA presented in Tables \ref{app:moment_results} and \ref{app:ViT_results} reflect the accuracy obtained for each configuration of these two hyperparameters, allowing us to explore the impact of different settings on performance and to choose the best hyperprameters to present the results in Table \ref{tab:moment_vit_results}. This flexibility in the PCA configuration allows us to adapt the method to a wide range of tasks, optimizing both performance and computational efficiency.

\setlength{\tabcolsep}{0.3em}
\begin{table}[h!]
\centering
\caption{Performance comparison between fine-tuning methods with different adapter configurations for the \texttt{MOMENT} foundation model} 
\label{app:moment_results} 
\scalebox{0.73}{
    \begin{tabular}{c|cccc}
    \toprule
    \multicolumn{1}{c}{\multirow{2}{*}{Dataset}} & 
    \multicolumn{4}{c}{adapter+head} \\
    \cmidrule(r{10pt}l{5pt}){2-5}
    & PCA & Scaled PCA & Patch\_8 & Patch\_16 \\
    
    \midrule
    DuckDuckGeese & 
    $0.667_{\pm 0.012}$ & 
    $0.533_{\pm 0.031}$ & 
    $0.567_{\pm 0.031}$ & 
    $0.573_{\pm 0.031}$ \\
    
    FaceDetection& 
    $0.566_{\pm 0.001}$ & 
    COM & 
    $0.582_{\pm 0.003}$ & 
    $0.558_{\pm 0.004}$ \\
    
    FingerMovement & 
    $0.573_{\pm 0.012}$ & 
    $0.563_{\pm 0.032}$ & 
    $0.633_{\pm 0.012}$ & 
    $0.563_{\pm 0.015}$ \\
    
    HandMovementDirection  & 
    $0.365_{\pm 0.036}$ & 
    $0.356_{\pm 0.043}$ & 
    $0.464_{\pm 0.021}$ & 
    $0.383_{\pm 0.021}$ \\
    
    Heartbeat & 
    $0.732_{\pm 0.005}$ & 
    $0.728_{\pm 0.003}$ & 
    $0.738_{\pm 0.007}$ & 
    $0.741_{\pm 0.013}$ \\
    
    InsectWingbeat & 
    $0.224_{\pm 0.003}$ & 
    $0.239_{\pm 0.003}$ & 
    $0.458_{\pm 0.002}$ & 
    $0.459_{\pm 0.004}$ \\
    
    JapaneseVowels & 
    $0.803_{\pm 0.003}$ & 
    $0.723_{\pm 0.020}$ & 
    $0.967_{\pm 0.002}$ & 
    $0.963_{\pm 0.002}$ \\
    
    MotorImagery & 
    $0.607_{\pm 0.012}$ & 
    $0.590_{\pm 0.020}$ & 
    $0.577_{\pm 0.006}$ & 
    $0.597_{\pm 0.015}$ \\
    
    NATOPS & 
    $0.739_{\pm 0.017}$ & 
    $0.731_{\pm 0.012}$ & 
    $0.857_{\pm 0.003}$ & 
    $0.915_{\pm 0.003}$ \\
    
    PEMS-SF & 
    $0.511_{\pm 0.022}$ & 
    $0.678_{\pm 0.007}$ & 
    $0.719_{\pm 0.012}$ & 
    $0.696_{\pm 0.018}$ \\
    
    PhonemeSpectra & 
    $0.212_{\pm 0.002}$ & 
    $0.227_{\pm 0.008}$ & 
    $0.224_{\pm 0.001}$ & 
    $0.186_{\pm 0.001}$ \\
    
    SpokenArabicDigits & 
    $0.978_{\pm 0.000}$ & 
    $0.963_{\pm 0.001}$ & 
    $0.967_{\pm 0.001}$ & 
    $0.956_{\pm 0.001}$ \\
    
    \bottomrule
    \end{tabular}
}
\end{table}

\setlength{\tabcolsep}{0.3em}
\begin{table}[h!]
\centering
\caption{Performance comparison between fine tuning methods with different adapter configurations for \texttt{ViT} foundation model} 
\label{app:ViT_results} 
\scalebox{0.73}{
    \begin{tabular}{c||cccc}
    \toprule
    \multicolumn{1}{c}{\multirow{2}{*}{Dataset}} & 
    \multicolumn{4}{c}{adapter+head} \\
    \cmidrule(r{10pt}l{5pt}){2-5}
    & PCA & Scaled PCA & Patch\_8 & Patch\_16 \\
    
    \midrule
    DuckDuckGeese & 
    $0.558_{\pm 0.023}$ & 
    $0.522_{\pm 0.023}$ & 
    $0.467_{\pm 0.031}$ & 
    $0.440_{\pm 0.035}$ \\
    
    FaceDetection & 
    $0.554_{\pm 0.001}$ & 
    $0.550_{\pm 0.010}$ & 
    $0.551_{\pm 0.003}$ & 
    $0.547_{\pm 0.007}$ \\
    
    FingerMovement & 
    $0.593_{\pm 0.044}$ & 
    $0.583_{\pm 0.023}$ & 
    $0.530_{\pm 0.036}$ & 
    $0.570_{\pm 0.053}$ \\
    
    HandMovementDirection  & 
    $0.367_{\pm 0.042}$ & 
    $0.327_{\pm 0.056}$ & 
    $0.396_{\pm 0.021}$ & 
    $0.369_{\pm 0.021}$ \\
    
    Heartbeat & 
    $0.736_{\pm 0.010}$ & 
    $0.734_{\pm 0.014}$ & 
    $0.766_{\pm 0.005}$ & 
    $0.763_{\pm 0.018}$ \\
    
    InsectWingbeat & 
    $0.344_{\pm 0.013}$ & 
    $0.268_{\pm 0.005}$ & 
    $0.287_{\pm 0.011}$ & 
    $0.266_{\pm 0.006}$ \\
    
    JapaneseVowels & 
    $0.890_{\pm 0.008}$ & 
    $0.865_{\pm 0.016}$ & 
    $0.922_{\pm 0.009}$ & 
    $0.921_{\pm 0.011}$ \\
    
    MotorImagery & 
    $0.567_{\pm 0.006}$ & 
    $0.552_{\pm 0.045}$ & 
    $0.593_{\pm 0.025}$ & 
    $0.573_{\pm 0.065}$ \\
    
    NATOPS & 
    $0.837_{\pm 0.012}$ & 
    $0.840_{\pm 0.017}$ & 
    $0.874_{\pm 0.014}$ & 
    $0.870_{\pm 0.008}$ \\
    
    PEMS-SF & 
    $0.584_{\pm 0.010}$ & 
    $0.613_{\pm 0.025}$ & 
    $0.634_{\pm 0.013}$ & 
    $0.674_{\pm 0.032}$ \\
    
    PhonemeSpectra & 
    $0.270_{\pm 0.003}$ & 
    $0.262_{\pm 0.008}$ & 
    $0.234_{\pm 0.002}$ & 
    $0.205_{\pm 0.006}$ \\
    
    SpokenArabicDigits & 
    $0.962_{\pm 0.003}$ & 
    $0.952_{\pm 0.003}$ & 
    $0.921_{\pm 0.006}$ & 
    $0.899_{\pm 0.002}$ \\
    
    \bottomrule
    \end{tabular}
}
\end{table}

\subsection{lcomb's Hyperparameter Sensitivity}

\begin{figure}[h!]
    \centering
    \begin{subfigure}[b]{0.49\textwidth}
        \centering
        \includegraphics[scale=0.27]{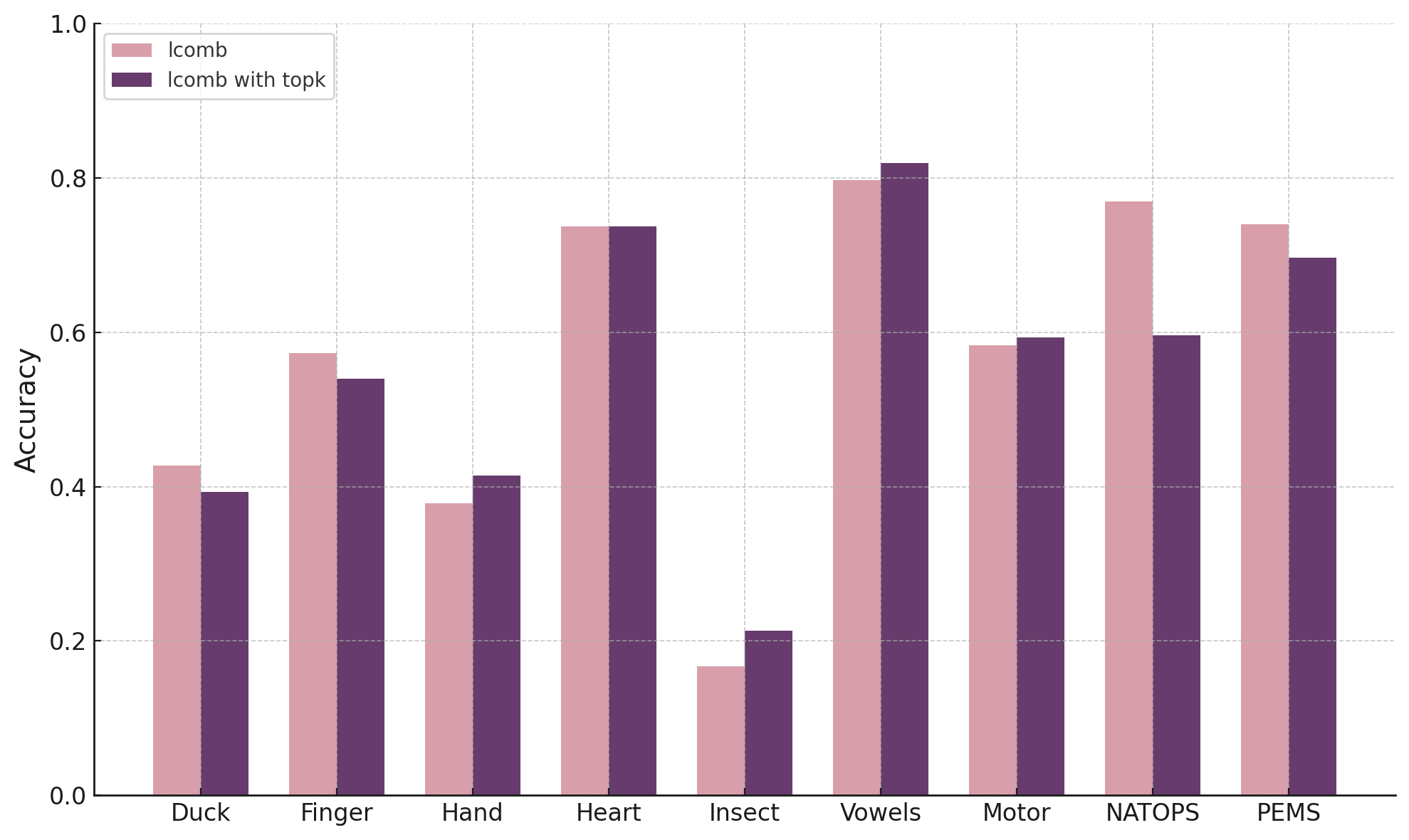}  
        \caption{\texttt{MOMENT}}
        \label{fig:moment_lcomb_full}
    \end{subfigure}%
    \hfill
    \begin{subfigure}[b]{0.49\textwidth}
        \centering
        \includegraphics[scale=0.27]{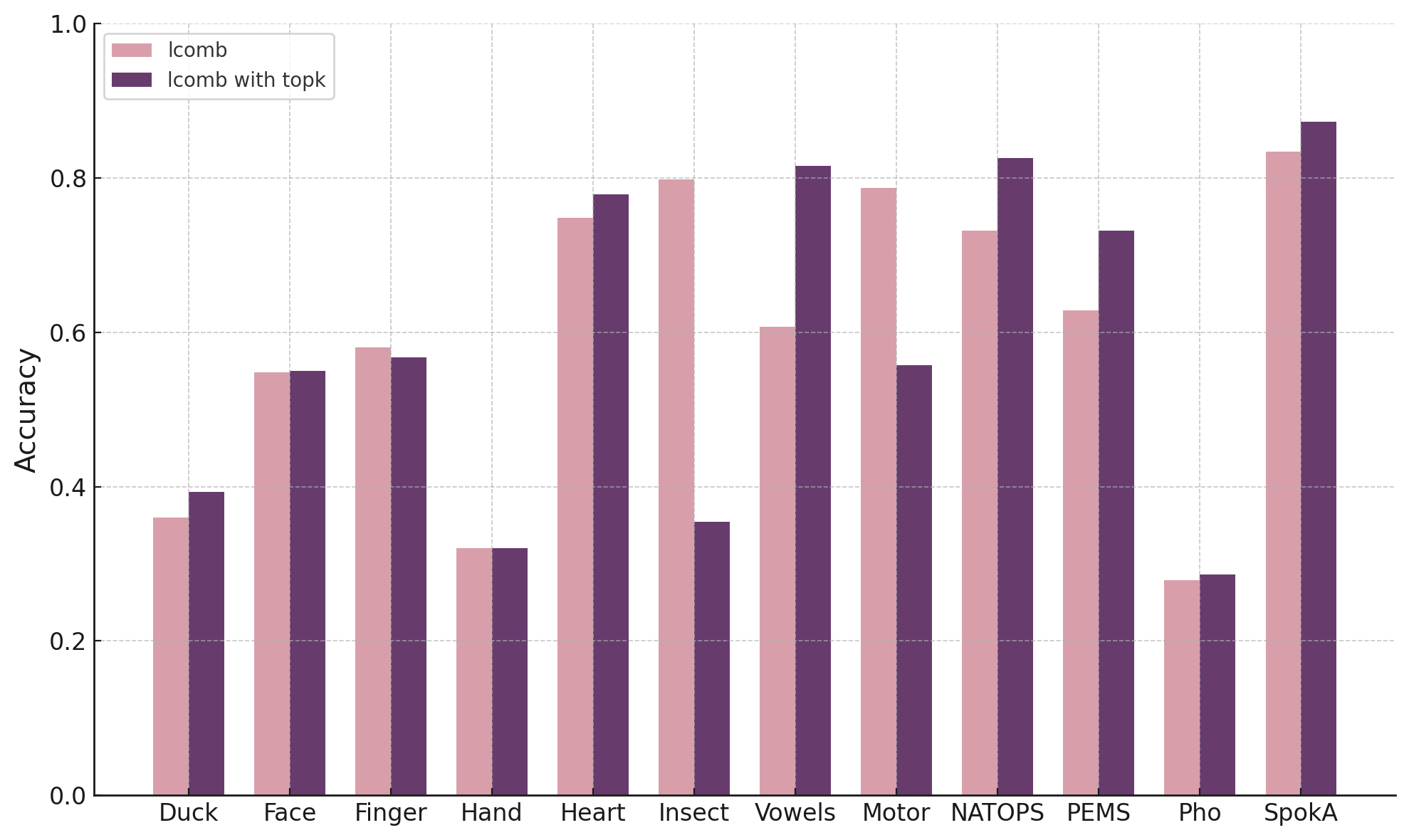}  
        \caption{\texttt{ViT}}
        \label{fig:ViT_lcomb_full}
    \end{subfigure}
    \caption{Performance Comparison Between \textit{lcomb} and \textit{lcomb\_top\_k} Fine-Tuning Configurations for both \texttt{MOMENT} and \texttt{ViT} Models}

    \label{fig:lcomb-adapter+head}
\end{figure}

In addition to the standard \textit{lcomb} configuration, we evaluated a variant called \textit{lcomb\_top\_k}, which introduces a form of regularization to make the attention mechanism more stable. In \textit{lcomb\_top\_k}, only the top $k$ largest attention weights are selected, and each row of the attention matrix is rescaled by dividing by the sum of these $k$ weights. For our experiments, we set $k = 7$. This mechanism is designed to reduce noise in the attention distribution, focusing the model on the most important relationships between elements in the input. The results shown in Figure \ref{fig:lcomb-adapter+head} show the performance comparison between \textit{lcomb} and \textit{lcomb\_top\_k} across several datasets for both MOMENT and \texttt{ViT} foundation models.


\subsection{Rank Comparisons}

\begin{figure}[h!]
    \centering
    \begin{subfigure}[b]{0.48\textwidth}
        \centering
        \includegraphics[scale=0.25]{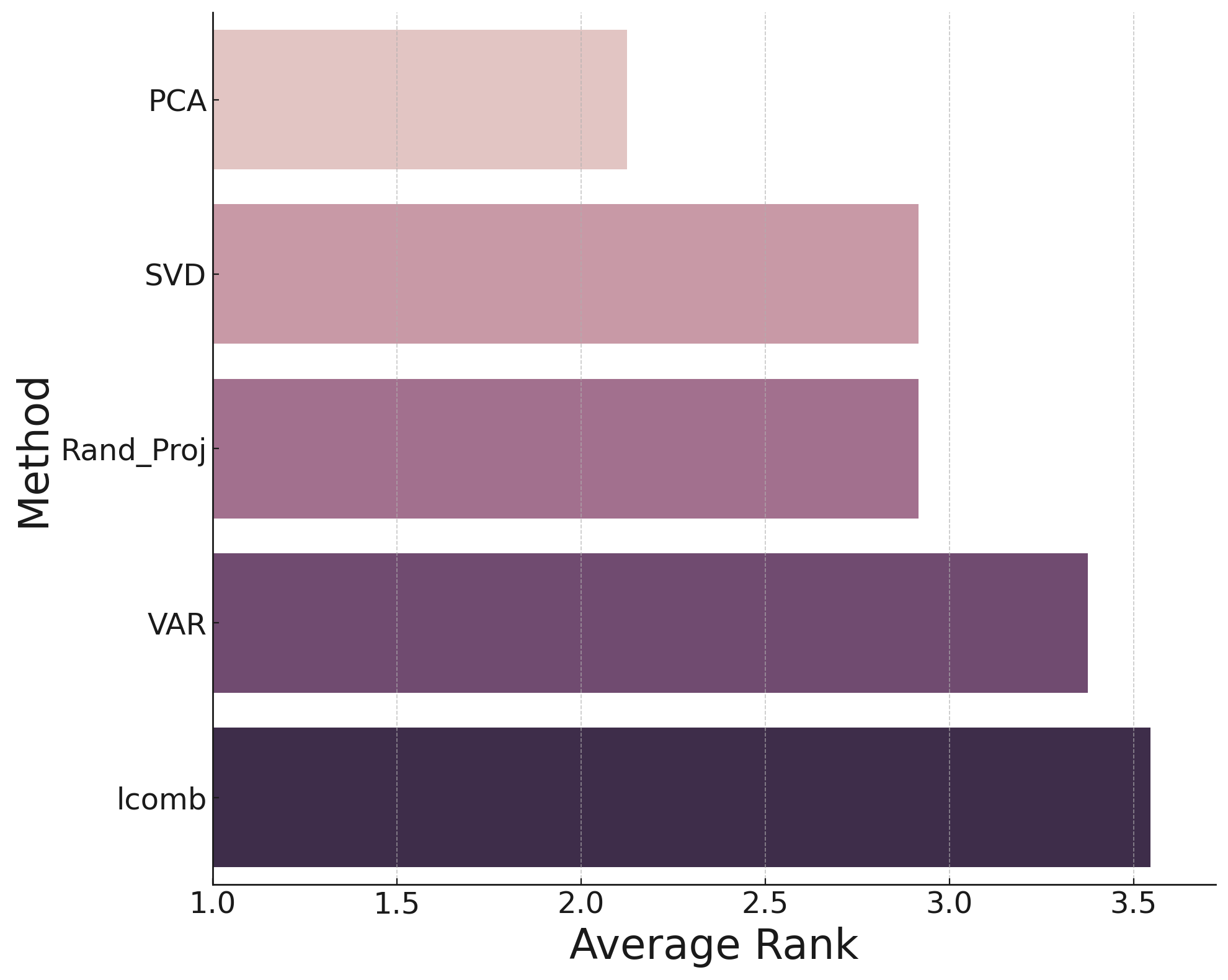}  
        \caption{Adapter's Average Rank for \texttt{MOMENT} Foundation Model}
        \label{fig:moment_rank}
    \end{subfigure}%
    \hfill
    \begin{subfigure}[b]{0.48\textwidth}
        \centering
        \includegraphics[scale=0.25]{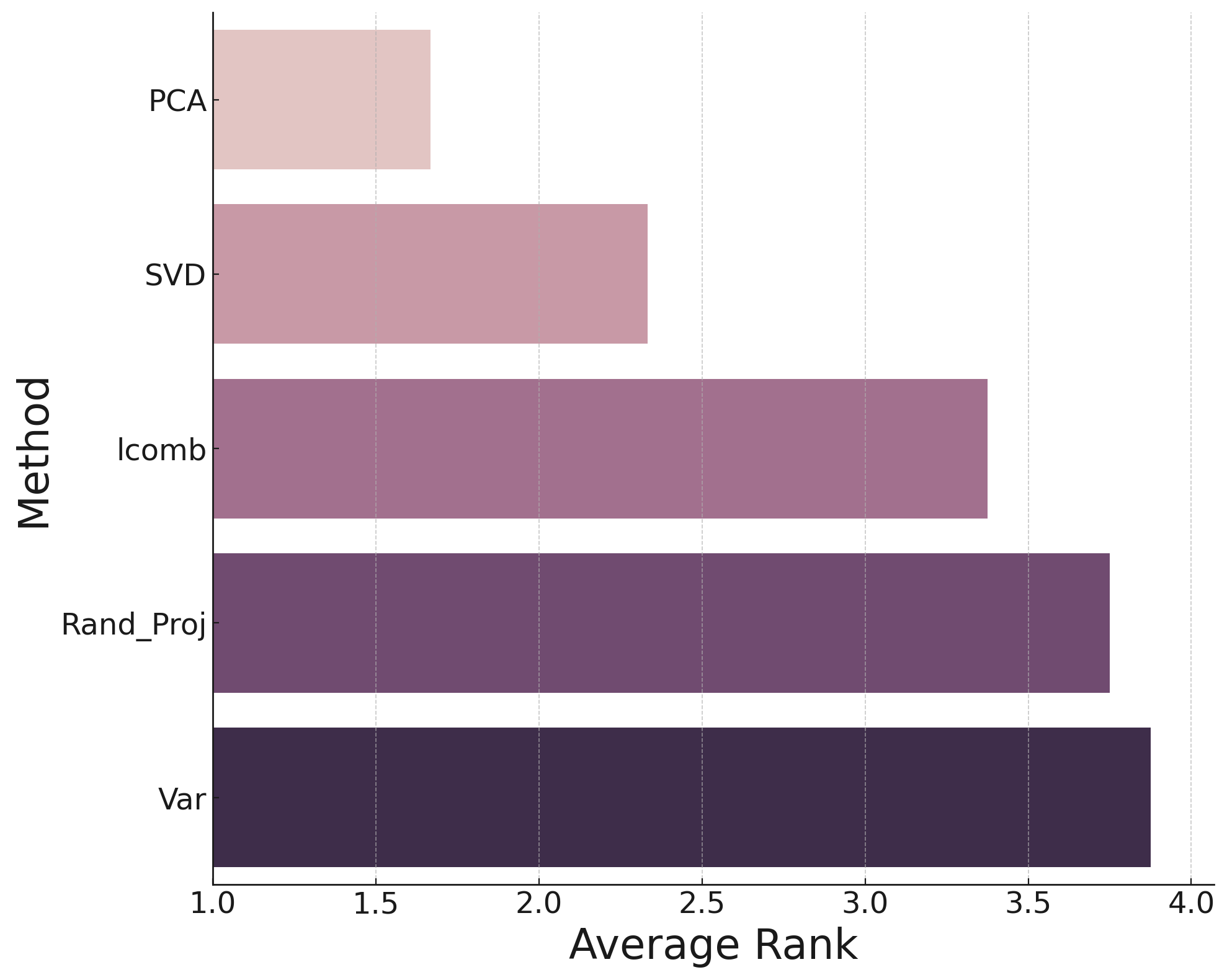}  
        \caption{Adapter's Average Rank for \texttt{ViT} Foundation Model}
        \label{fig:ViT_rank}
    \end{subfigure}
    \caption{Comparison of Adapter's Average Rank for \texttt{MOMENT} and \texttt{ViT} Foundation Models averaged across all datasets and three different seeds}
    \label{fig:comparison_ViT_moment_rank}
\end{figure}

Figure \ref{fig:comparison_ViT_moment_rank} shows a comparison of the average rank for different adapter methods used in the \texttt{MOMENT} and \texttt{ViT} foundation models. The average ranks were computed across all datasets and averaged over three seeds. The comparison gives insight into the relative performance of each adapter method when applied to these two models.

For the \texttt{MOMENT} foundation model, as depicted in Figure \ref{fig:moment_rank}, the \textit{PCA} adapter ranks the lowest, indicating the best performance, while the \textit{lcomb} adapter ranks the highest, showing relatively lower performance. The remaining adapters—\textit{SVD}, \textit{Rand\_Proj}, and \textit{VAR}—lie in between, with \textit{Rand\_Proj} and \textit{SVD} showing close performance.

Similarly, in the case of the \texttt{ViT} foundation model (Figure \ref{fig:ViT_rank}), \textit{PCA} exhibits the lowest average rank, implying superior performance. \textit{Rand\_Proj} also performs relatively worse in this case. The consistency of PCA's superior performance across both models highlights its effectiveness

\subsection{Statistical Tests}
\label{app:stats}

\begin{figure}[h!]
    \centering
    \begin{subfigure}[b]{0.48\textwidth}
        \centering
        \includegraphics[scale=0.28]{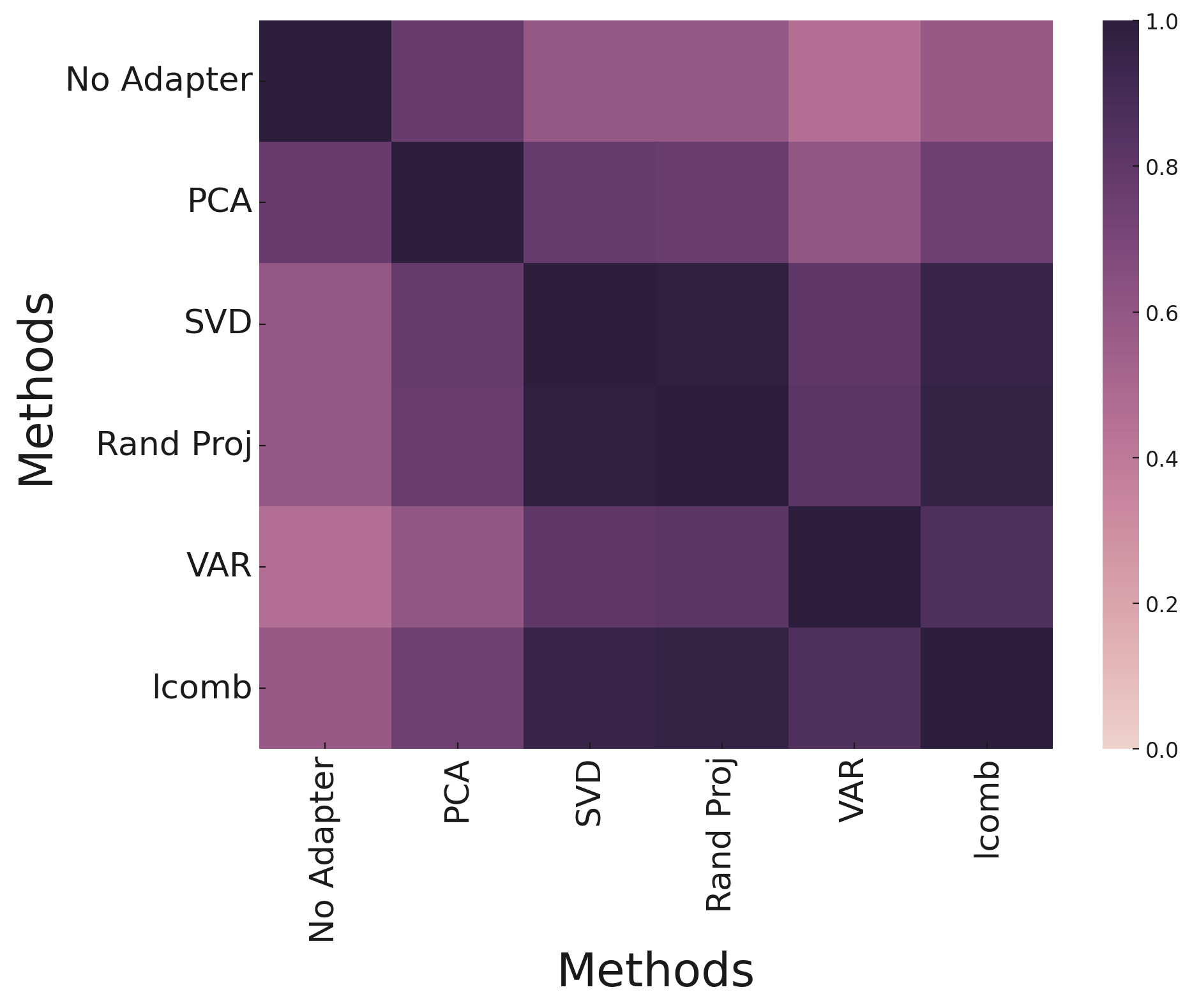}  
        \caption{Heatmap of Pairwise p-values for Adapter Methods for \texttt{MOMENT} Foundation Model}
        \label{fig:moment_running_time}
    \end{subfigure}%
    \hfill
    \begin{subfigure}[b]{0.48\textwidth}
        \centering
        \includegraphics[scale=0.28]{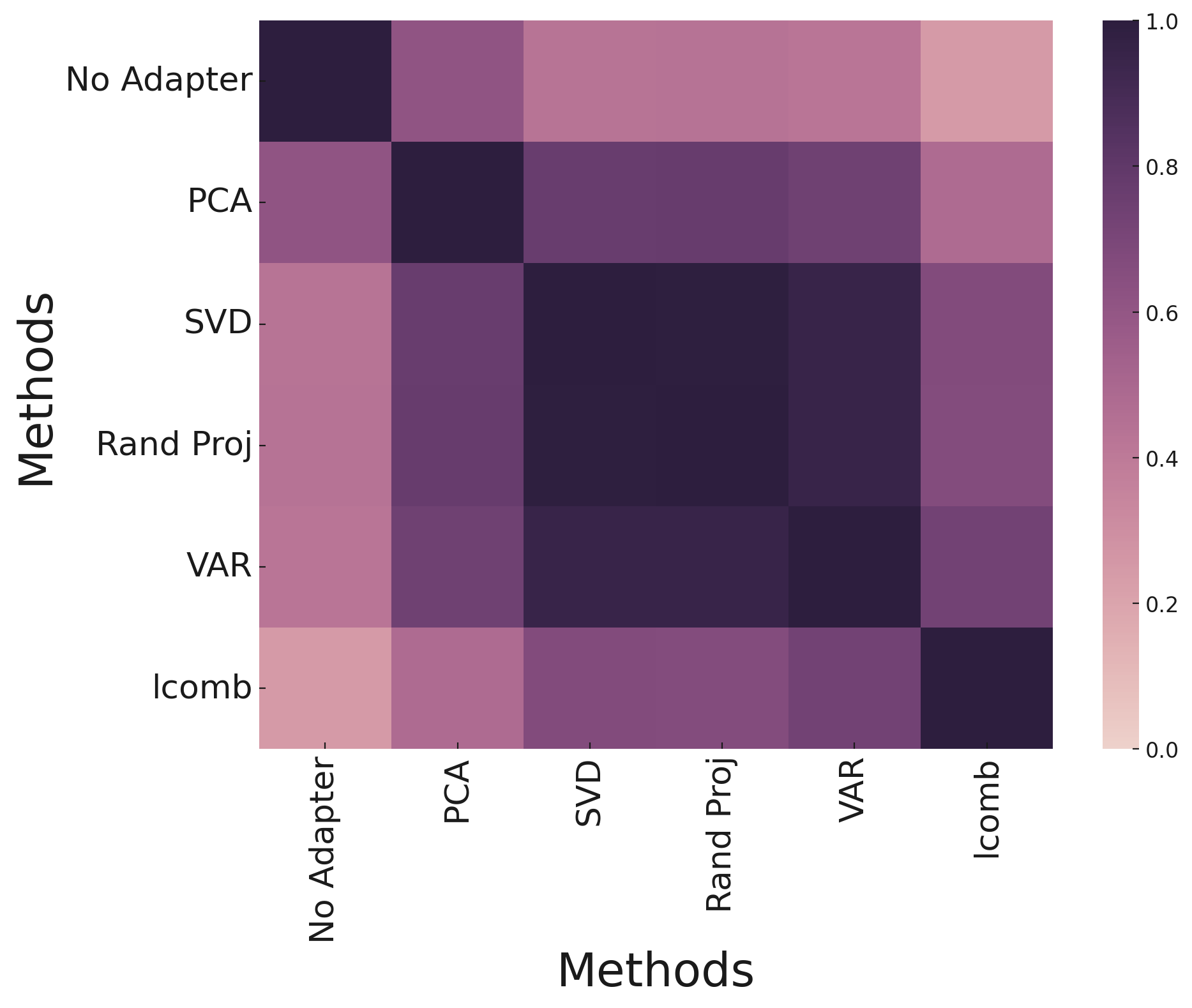}  
        \caption{Heatmap of Pairwise p-values for Adapter Methods for \texttt{ViT} Foundation Model}
        \label{fig:ViT_running_time}
    \end{subfigure}
    \caption{Heatmap of Pairwise p-values for Adapter Methods for \texttt{MOMENT} and \texttt{ViT} Foundation Models averaged across all datasets and three different seeds}
    \label{fig:heatmap_ViT_moment}
\end{figure}

The heatmap shown in Fig. \ref{fig:heatmap_ViT_moment} present the pairwise p-values between different fine-tuning methods applied to the \texttt{MOMENT} and \texttt{ViT} foundation models across several datasets. The methods compared include \textit{No Adapter}, \textit{PCA}, \textit{SVD}, \textit{Rand Proj}, \textit{VAR}, and \textit{lcomb}. The p-values were calculated using a two-sample Student's t-test with unequal variances, based on accuracy results obtained from three different seeds for each method.

The null hypothesis for each comparison states that there is no significant difference in the mean performance, in terms of accuracy, between the two methods being compared. A p-value close to 1 supports this hypothesis, indicating that the two methods yield statistically similar performance. In contrast, a p-value close to 0 suggests a significant difference.  In the \texttt{MOMENT} heatmap, the lowest p-value observed is $0.46$, while for ViT, the minimum p-value is $0.25$. These visualizations indicate that there is no statistically significant difference between fine-tuning using adapter + head with different adapters, and similarly, no difference is observed between adapter + head and head-only fine-tuning, regardless of the adapter used.

\subsection{Full Fine-Tuning Regime}
\label{app:full}

\begin{figure}[h!]
    \centering
    \begin{subfigure}[b]{0.48\textwidth}
        \centering
        \includegraphics[scale=0.3]{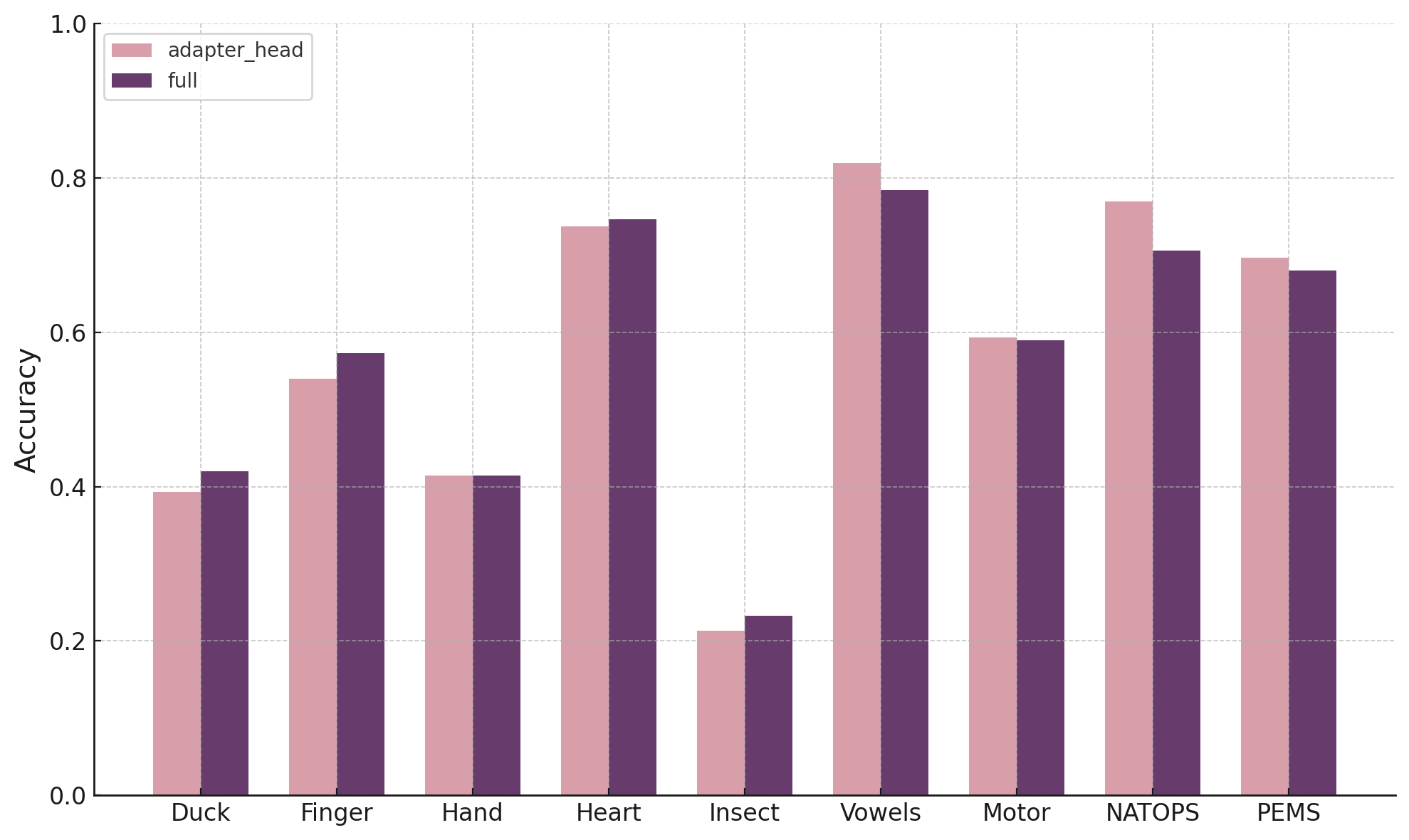}  
        \caption{\texttt{MOMENT}}
        \label{fig:moment_lcomb_full}
    \end{subfigure}%
    \hfill
    \begin{subfigure}[b]{0.48\textwidth}
        \centering
        \includegraphics[scale=0.3]{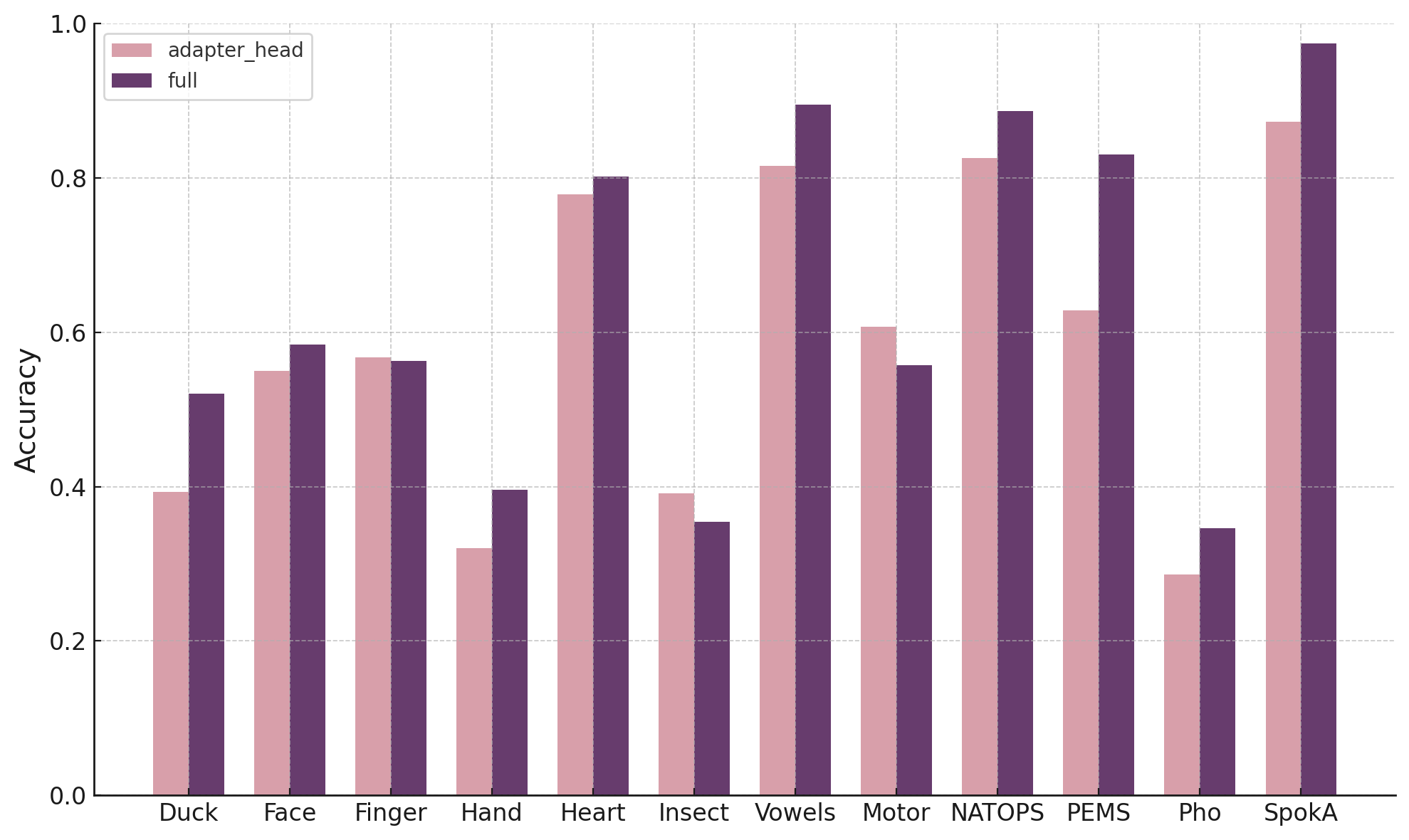}  
        \caption{\texttt{ViT}}
        \label{fig:ViT_lcomb_full}
    \end{subfigure}
    \caption{Full fine-tuning vs tuning adapter+head for \texttt{lcomb}.}
    \label{fig:lcomb-full}
\end{figure}

\end{document}